\documentclass[conference]{IEEEtran}
\IEEEoverridecommandlockouts
\usepackage{cite}
\usepackage{amsmath,amssymb}
\usepackage{graphicx}
\usepackage{booktabs}
\usepackage{url}
\usepackage[hidelinks]{hyperref}
\usepackage{microtype}
\newcommand{\nPlanted}{545}
\newcommand{\nSingle}{169}
\newcommand{\nConjunctive}{126}
\newcommand{\nOverdetermined}{70}
\newcommand{\nAgentStratum}{180}
\newcommand{\nPolicies}{3}
\newcommand{\agentOnlyInfra}{0}
\newcommand{\agentOnlyVerdict}{33.0}
\newcommand{\agentOnlyMisattr}{100}
\newcommand{\agentOnlyDollars}{114,383.40}
\newcommand{\agentOnlyRepair}{0.0}
\newcommand{\faultSingleOverdet}{0}
\newcommand{\faultSingleRepair}{82.4}

\newcommand{\minSuffRepair}{100.0}
\newcommand{\minSuffReplays}{2.3}
\newcommand{\shapleyReplays}{280.1}

\newcommand{\dollarsAvailable}{82,488.60}

\newcommand{\overdetRate}{5.6}
\newcommand{\interactionGap}{76.73}
\newcommand{\shapleyError}{1.88}
\newcommand{\meanInstances}{7.6}
\newcommand{\maxInstances}{12}
\newcommand{\decompExact}{3852/3852}
\newcommand{\reactInfra}{59.3}
\newcommand{\reactPolicy}{17.3}
\newcommand{\ruleInfra}{75.1}
\newcommand{\optInfra}{-6.8}

\newcommand{\gateIrreducible}{100.0}
\newcommand{\gateMeanLoss}{31.37}
\newcommand{\negPolicyRate}{6.9}
\newcommand{\negPolicyWorst}{-1,442.60}
\newcommand{\superEpisodes}{240}
\newcommand{\superFailed}{199}
\newcommand{\superFailRate}{82.9}
\newcommand{\superCurrentOk}{240}
\newcommand{\exOrder}{24.99}
\newcommand{\exLoss}{28.00}
\newcommand{\exInstances}{7}
\newcommand{\exShare}{14.00}
\newcommand{\exZeroed}{5}
\newcommand{\exAgentOnly}{28.00}

\newcommand{\nonAdditiveCI}{27.8\% (95\% CI: 23.3--32.3\%)}
\newcommand{\overdetCI}{5.6\% (95\% CI: 3.1--8.7\%)}
\newcommand{\ciTasks}{219}
\newcommand{\ciReplicates}{2,000}

\newcommand{\ciEpisodes}{320}
\newcommand{\searchCap}{3}
\newcommand{\largestRepairSet}{2}
\newcommand{\capBound}{0}

\begin{document}

\title{CausalLoss-Fin: Attributing Financial-Agent Loss to
Decisions and Infrastructure Faults}

\author{\IEEEauthorblockN{Abhishek Sharma, \textit{Senior Member, IEEE}}
\IEEEauthorblockA{\texttt{abhicse24@gmail.com} \quad ORCID: 0009-0007-1103-2103}}

\maketitle

\begin{abstract}
When an agent handling a payment exception loses money, the agent-step
attribution methods this paper compares against will name one of its actions.
They will do so even when a settlement message was dropped and the agent never
had a chance: they intervene on agent actions and do not expose infrastructure
faults as intervenable variables, so every dollar they explain is charged to a
decision.

We take a benchmark whose fault process is explicit and replayable, decompose
each episode's realised delivery schedule into named, individually repairable
messages, and intervene on \emph{both} the agent's choices and the
infrastructure's. A telescoping identity splits any policy's loss exactly three
ways: an infrastructure effect, a policy differential against the best
implementable policy, and a reference-policy residual. Two of the three can be negative, so none is a share; Shapley then divides the
first into signed allocations over individual messages.

One result is structural and needs no corpus: an agent-only baseline identifies
no infrastructure cause, because its model contains no variable that could name
one. What \nPlanted{} planted episodes across \nPolicies{} policies measure is
the size of that consequence. It misfiles \agentOnlyMisattr\%\ of
infrastructure episodes and charges \$\agentOnlyDollars{} to the agent.
Repairing what it names recovers \agentOnlyRepair\%\ of the available loss;
repairing a minimal sufficient set recovers \minSuffRepair\%. Scoring messages
one at a time is not merely imprecise: \nonAdditiveCI{} of episodes do not
decompose additively.

We evaluate deterministic programmatic policies rather than language-model
agents, which is what makes replay exact and which limits external validity to
stochastic agents. The prevalence figures are properties of this generator, not
field rates.
\end{abstract}


\section{Introduction}

An agent is asked to resolve a payment exception. It reads the processor, which
says the capture is still pending, waits, reads again, sees nothing, and closes
the case. The capture had in fact settled four minutes earlier; the message
saying so was dropped in transit. The customer is charged and never receives
the goods, and the loss is the order value plus a dispute fee.

Now ask the question every incident review asks: whose fault was that?

Recent work answers it by intervention. Causal Agent Replay~\cite{car} models a
run as a structural causal model, applies a \emph{do} operation to a step,
re-executes the trajectory forward under the same policy, and reports the shift
in the outcome distribution. CausalFlow~\cite{causalflow} computes
step-level causal responsibility scores and then synthesises minimal repairs.
REFLECT~\cite{reflect} diagnoses candidate error steps, replays with
diagnosis-specific patches, and uses the verified outcome flip as contrastive
evidence. These are careful methods and a clear advance on judging traces with
a language model, which is correlational and, on one published benchmark,
around 14\% accurate at the step level~\cite{car}.

They also share an assumption. Each intervenes on the agent's steps and
exposes no infrastructure fault as an intervenable variable. In an environment
where the infrastructure is itself a causal source --- messages delayed,
duplicated, dropped, reordered --- a causal model with no infrastructure
variable cannot name an infrastructure cause. Asked to
explain the episode above, the best it can do is nominate the step that acted
on information the agent never received. That is not a defect of execution. It
is a property of the model, and it is the kind of thing that is easy to miss
until the environment is one where the answer is checkable.

This paper makes it checkable. We build on FinalityBench~\cite{finalitybench},
an executable benchmark in which four financial systems (payment processor, double-entry ledger, ERP,
bank feed) are fed independently faulted delivery streams derived from a hidden canonical event log, and agents are graded on
executed monetary effect. Its fault process is explicit, seeded and replayable,
which is what makes the counterfactual ``what if that message had arrived''
executable rather than hypothetical.

Our contributions:

\begin{itemize}
\item \textbf{Joint intervention.} We intervene on agent decisions and on
individual infrastructure faults in the same causal model, on the same episode,
with exact common random numbers (Section~\ref{sec:scm}).

\item \textbf{An exact three-way split.} A telescoping identity decomposes any
policy's loss into an infrastructure effect, a policy differential and a
reference-policy residual, with nothing left unattributed. It holds on
\decompExact{} episodes checked (Section~\ref{sec:split}).

\item \textbf{A decomposition of the fault process into repairable units},
together with the finding that FinalityBench's fault families compose instead of superposing, so per-family
attribution is unsound, and we say why (Section~\ref{sec:faults}).

\item \textbf{Planted ground truth in four strata}, including the
overdetermined case that single-effect scoring provably cannot handle, with
verified harmless distractors so that naming a cause is not trivial
(Section~\ref{sec:planted}).

\item \textbf{A measurement of what agent-only attribution costs} when the
environment is at fault, in identification accuracy, in allocation error, in
dollars, and in loss that repairing the named cause fails to recover
(Section~\ref{sec:results}).
\end{itemize}

As in the benchmark this builds on, no language model was evaluated: no model
endpoint was available. The policies whose failures we attribute are
deterministic procedures, which sharpens the causal analysis, interventions are
exact rather than sampled, and narrows what the results say about agents that
reason. Section~\ref{sec:limits} is explicit about the difference.

\section{Related work}

\textbf{Counterfactual attribution for agent failures.} CAR~\cite{car} is the
closest work and the one we compare against. It executes same-policy
interventions rather than substituting an oracle or asking a model to judge
text, reports distributional outcomes with confidence intervals, and includes
Shapley credit-splitting and ground-truth validation. CausalFlow~\cite{causalflow}
adds counterfactual repair, generating minimally edited traces that flip the
outcome. REFLECT~\cite{reflect} targets silent failures, where the agent
completes the task and the result is wrong with no error signal. All three intervene on agent steps and expose no infrastructure fault as an
intervenable variable, which we checked in each paper directly. Our
disagreement with this line is narrow. The machinery is right; the variable
set is incomplete for environments where infrastructure causes loss, and
Section~\ref{sec:results} quantifies what that incompleteness costs.

\textbf{Attribution that does intervene on the environment.} It would be wrong
to say that no prior work perturbs anything but the agent. AgenticRAG-
FP~\cite{agenticragfp} injects certified faults at specified retrieval hops of
a multi-hop RAG pipeline, corrupting answer-bearing and bridge facts while
leaving documents topically intact, and re-executes the downstream trajectory
to ask whether a post-hoc trace still identifies the injected hop. That is an
environment intervention and it is the closest prior work to ours in that
respect. What it measures is diagnostic accuracy --- was the injected hop
recovered --- rather than how much of a realised cost each fault accounts for,
and it has no counterpart to the signed decomposition below.
Causely~\cite{causely} goes the other way, giving agents a structured causal
representation of environment topology and dependencies so that they diagnose
incidents better; the causal structure is an input to the agent rather than an
instrument for apportioning loss after the fact.

\textbf{Attribution beyond a single cause.} DCFA~\cite{dcfa} attributes
failure in multi-agent systems by combining a dependency graph over traces
with counterfactual refinement, and reports step-level accuracy; its
interventions are over agent interactions. MP-Bench~\cite{mpbench} argues that
a multi-agent failure often admits several plausible attributions and that
treating one as uniquely correct is an artifact of benchmark design. Our
overdetermined stratum is the same phenomenon in a setting where the competing
explanations are messages rather than agents, and the signed decomposition is
one answer to it: when two faults are each sufficient, Shapley splits the
credit evenly instead of forcing a choice between them.

\textbf{Actual causation.} The question of which of several events counts as
the cause of an outcome is the subject of the structural-model account of
actual causality~\cite{halpern2016}, in which a cause is a but-for cause
relative to a contingency, and of the associated notion of degree of
responsibility~\cite{chockler2004}. Overdetermination, two events, either
sufficient alone, so neither is necessary, is the standard hard case, and it is
the one our third stratum is built from. We take the practical route: a minimal sufficient set is the smallest set of
messages whose joint repair removes the loss. The search stops at \searchCap{} messages, which would make a perfect
recovery figure conditional on the cases small enough to fit inside that
bound. It does not: over \ciEpisodes{} episodes the largest set found holds
\largestRepairSet{} messages and the cap binds on \capBound{} of them, so the
recovery figure is a property of the method and not of the cap.

\textbf{Shapley allocation.} Distributing a total among contributors so that
the parts sum to the whole is the Shapley value~\cite{shapley1953}. It has been
carried into root-cause analysis, though not without warnings: Kelen et
al.~\cite{shapleyrca} show that the \emph{asymmetric} variant, which relaxes the
symmetry axiom to encode a causal ordering, produces counter-intuitive
attributions outside a restricted model class. We use the standard symmetric
value, and for one specific job: dividing the infrastructure effect across
individual messages in a way that survives interaction. The infrastructure effect itself is not estimated by Shapley; it comes from
the exact identity, so the efficiency axiom is a check on the implementation
rather than a modelling assumption.

\textbf{The environment.} FinalityBench~\cite{finalitybench} supplies the
setting: a hidden canonical event log, four projections fed by separately
faulted delivery streams, an effect-level monetary oracle, and paired
counterfactual tasks. We consume its released
artifact (\textsc{doi} 10.5281/zenodo.22262591) rather than reimplementing it,
and every result here records which revision it ran against.

\section{The setting}

A FinalityBench episode is one order. A capture is submitted, and its terminal
outcome, settled or failed, becomes a fact at some later instant that no tool
can reveal early. Four systems learn about events through delivery streams that
a seeded fault engine perturbs in six ways: delay, duplication, loss,
reordering, half-committed ledger postings, and stale reads. An agent has
eleven tools, five of them irreversible, and a ship-by deadline. Grading is on
the merchant's terminal economic position, reported as the shortfall against a
privileged reference that is told when the capture resolves.

The property we need is that an episode is a pure function of the case, the
seed, the policy and the delivery schedule. Replay with one message restored is
therefore exact: two worlds that differ in one repaired delivery differ in
nothing else, and no averaging over noise is required to see the effect.

\section{Fault instances}
\label{sec:faults}

Attribution needs a cause one can point at. ``Delay was enabled'' is a knob;
``the settlement message to the processor never arrived'' is a cause. We
decompose a realised schedule into named instances by diffing it against the
fault-free schedule for the same case and seed: every way the two differ is an
instance, and repairing all of them returns the fault-free world exactly. There
are \meanInstances{} instances per episode on average and at most
\maxInstances. An archetype's defining fault, the settlement the processor
never hears about, appears as an instance too, flagged as structural, so it can be intervened on instead of being baked into the background.

A first version decomposed per family instead, building each single-family
schedule and diffing that against the baseline. Pooling those instances and
applying them to the fault-free schedule fails to reproduce the realised one on
\superFailed{} of \superEpisodes{} episodes (\superFailRate\%), where the
decomposition described above reproduces all \superCurrentOk{}. The reason is
a property of the environment rather than a coding slip. FinalityBench draws its fault families
independently but they do not \emph{act} independently: duplicate copies are
placed relative to a delivery's already-delayed arrival, a half-commit
propagates to copies that exist only because duplication fired, and whether a
reorder swap is eligible at all depends on times that delay has already moved.
Families that compose cannot be attributed by superposition. The family label
therefore annotates an instance rather than defining it, and the label is assigned by a but-for test taken in context: remove that
family from the full profile and ask whether this delivery still arrives when
it did. Asked the
naive way, in isolation, reordering appears never to fire at all; asked in
context it accounts for a fifth of the moved deliveries, because the swaps it
makes only become possible once delay has bunched arrivals together.

\section{The causal model}
\label{sec:scm}

\subsection{Variables and structural equations}

An episode is generated by the following model. The exogenous variables are the
case $C$ (order value, when the capture resolves, whether it settles, whether a
retry would work) and the fault realisation $F = \{f_1,\dots,f_m\}$, the set of
individual message-level instances of Section~\ref{sec:faults}. The endogenous
variables are the delivery schedule, each system's view at each instant, the
agent's actions, the executed effects, and the loss:

\begin{align}
S      &= \textsc{schedule}(C, F) \\
V_t^{(j)} &= \textsc{fold}_j\bigl(\{d \in S : \textsc{arr}(d) \le t - \lambda_j\}\bigr) \\
A_k    &= \pi\bigl(H_{k-1}\bigr), \qquad H_k = H_{k-1} \cup \{(A_k, O_k)\} \\
E      &= \textsc{effects}(A_1,\dots,A_n) \\
L      &= R - \textsc{position}(C, E)
\end{align}

with $j$ ranging over the four systems, $\lambda_j$ that system's read lag, $\pi$
the policy, $H_k$ the history it has seen, and $R$ the reference position, held
fixed across every counterfactual so that a repaired message moves the policy's
position and nothing else.

Two things follow from the shape of these equations. First, $F$ enters only
through $S$ and thus only through the views $V^{(j)}$: in this environment the
infrastructure can harm an agent only by corrupting what it is able to see,
never by moving money directly. Second, every equation is deterministic, so
$do(f_i := \bot)$ and $do(A_k := a')$ are exact operations rather than
distributions to be sampled.

\subsection{Interventions}

Two interventions are available.

\emph{Repairing a message.} Rebuild the delivery schedule without one instance
and re-run the same policy on the same case. Because the schedule is an input
rather than a random draw, nothing else changes.

\emph{Substituting a decision.} Force a chosen action at step $k$ and let the
policy continue from there, seeing the consequences. We report the effect of a
step as the loss avoided by its best alternative, a regret, rather than as a
distribution shift, because the policies here are deterministic and resampling
under the same policy, the usual move, changes nothing at all.

\subsection{An exact three-way split}
\label{sec:split}

Write $w$ for the factual world, $w_0$ for the same case with every message
repaired, $\pi$ for the policy under study, and $\pi^{*}$ for the best policy
that uses no privileged information. Then

\begin{equation}
\begin{split}
L(\pi, w) = \underbrace{L(\pi,w) - L(\pi,w_0)}_{\text{infrastructure}}
          &+ \underbrace{L(\pi,w_0) - L(\pi^{*},w_0)}_{\text{policy}}\\
          &+ \underbrace{L(\pi^{*},w_0)}_{\text{irreducible}}
\end{split}
\end{equation}

The terms telescope, so the split is exact and there is no residual in which a
modelling error could hide.

Two of the three terms can be negative, so none of them is called a share.
They are the \emph{infrastructure effect}, the \emph{policy differential} and
the \emph{reference-policy residual}. The Shapley output is not called a share
either: it divides the infrastructure effect into parts that sum to it, but
those parts are signed, so we call them \emph{signed allocations} throughout. The infrastructure effect goes negative when faults help
a policy on net, which happens and is reported in Section~\ref{sec:results}. The policy term is a gap
to the best \emph{implementable} policy rather than to a per-task optimum, so it
goes negative whenever the subject beats that baseline on a task: on
\negPolicyRate\%\ of episodes, reaching \$\negPolicyWorst. Read the three
numbers as a signed decomposition that sums to the loss, not as a partition into
non-negative parts. This matters more than it might appear: a residual
term is precisely where unattributed loss would accumulate, and unattributed
loss in an agent-analysis tool gets read as the agent's fault. The identity holds on \decompExact{} episodes checked, and the test suite
asserts it. Being algebra, it is reported without an
interval: a confidence band around an identity would say nothing.

Shapley values then allocate the infrastructure effect across individual
messages, with coalition value $v(S) = L(\pi, w_0) - L(\pi, S)$. They satisfy
efficiency, so the allocations sum to the effect, and they are \emph{signed}:
a message whose repair would have increased the loss carries a negative
marginal contribution and therefore a negative allocation. Additivity of the
value function does not make them non-negative, and on this corpus the
infrastructure effect itself goes negative for one policy. Exact below
thirteen instances, permutation-sampled above; the sampled estimate sits within
\shapleyError\%\ of the exact one in $L_1$.

\section{Attribution methods}
\label{sec:methods}

Seven methods, sharing an interface: given an episode they return a ranked list
of candidate causes with scores in cents, and a verdict on which side is
responsible. Two are heuristics, one is the published shape we compare against,
and four use fault interventions.

\textbf{Last action.} Blame the final irreversible step. It is the reflex an
incident review starts from, and it is included because the observation that
motivates intervention-based attribution in the first place is that the step
which executes a harmful action is rarely the step that decided on
it~\cite{car}.

\textbf{First divergence.} Blame the earliest message that arrived late or not
at all. A plausible operations reflex, find the first thing that looks wrong
and call it the cause, which ignores entirely whether that message mattered.

\textbf{CAR-shaped agent-only baseline.} Substitute each of the agent's steps
in turn, run forward under the same policy, and rank the steps by loss
avoided.

It matters what this is and is not. It is not a reproduction of
CAR~\cite{car}, CausalFlow~\cite{causalflow} or REFLECT~\cite{reflect}, and no
number below should be read as a score for any of those systems. It is their
\emph{variable scope} --- interventions on agent steps only --- reduced to
essentials and transplanted here, with interventions exact rather than sampled because the policy is
deterministic, which gives the baseline favourable conditions on the dimension
evaluated here. Its
causal model contains no infrastructure variable, so all of its causes are
agent steps and its infrastructure effect is identically zero.

\textbf{Single-fault repair.} Repair each message on its own, rank by loss
avoided. The natural first thing to try once fault interventions are available,
and exactly right whenever the loss decomposes additively.

\textbf{Minimal sufficient set.} Search for the smallest set of messages whose
\emph{joint} repair removes the loss, smallest first. This is a but-for cause
taken as a set rather than a singleton, in the spirit of the structural-model
account of actual causation~\cite{halpern2016}: no member need be a cause on
its own, and together they are. It is the only method here that is both exact
on every stratum and cheap.

\textbf{Joint greedy.} Rank agent steps and messages together by their
individual effects. Included to show that combining the two sides naively is not enough: its
independently computed scores do not sum to the loss, because a step regret
and a message repair are not measured on the same scale.

\textbf{Joint Shapley.} Take the agent and infrastructure effects from the exact
identity of Section~\ref{sec:split}, then distribute the infrastructure effect
across individual messages by Shapley value, and rank within whichever side the
identity says is responsible. Ranking the two sides against each other by raw score does not work. Shapley
values sum to the infrastructure effect while step regrets do not live on that
scale, so a single step scored at the whole loss outranks two messages that
split it. An earlier version of this method lost every overdetermined episode
for exactly that reason.

\section{Planted causes}
\label{sec:planted}

Validating a method against causes the method itself derived proves nothing. We
build episodes the other way round: start from a case the policy solves
cleanly, inject a known set of faults, and check that the loss appears. The
injected set is then the cause by definition.

Four strata, each testing something different. \textbf{Single}
(\nSingle{} episodes): one fault causes the loss. \textbf{Conjunctive}
(\nConjunctive): two faults, neither harmful alone, harmful together. Each is necessary given
the other, so repairing either fixes the episode.
\textbf{Overdetermined} (\nOverdetermined): two faults, \emph{either} sufficient
on its own, so repairing one changes nothing and every one-at-a-time score is
zero. \textbf{Agent} (\nAgentStratum): no faults bite, and the policy loses
money anyway.

Every episode also carries distractor faults, verified harmless \emph{in the
presence of} the causal set rather than merely harmless alone. Without them the
task is trivial: an earlier version planted only the causal faults, and a
heuristic that named the earliest message in the schedule scored 100\%, because
there was nothing else to name.

\section{Results}
\label{sec:results}

\begin{table*}[t]
\centering
\caption{Identifying the planted cause. Verdict is the share of episodes on
which the method assigns responsibility to the correct side.}
\label{tab:accuracy}
\footnotesize\setlength{\tabcolsep}{3pt}
\begin{tabular}{lrrrrr}
\toprule
& \multicolumn{4}{c}{top-1 identification (\%)} & \\
\cmidrule(lr){2-5}
method & single & conjunctive & overdet. & agent & verdict (\%) \\
\midrule
last action & 0 & 0 & 0 & 100 & 33.0 \\
first divergence & 7 & 13 & 53 & 0 & 67.0 \\
agent-only intervention & 0 & 0 & 0 & 100 & 33.0 \\
single-fault repair & 100 & 100 & 0 & 100 & 100.0 \\
\bfseries joint, greedy & 100 & 100 & 0 & 100 & 100.0 \\
\bfseries minimal sufficient set & 100 & 100 & 100 & 100 & 100.0 \\
\bfseries joint Shapley & 100 & 100 & 100 & 100 & 100.0 \\
\bottomrule
\end{tabular}
\end{table*}

\begin{figure}[t]
\centering
\includegraphics[width=\columnwidth]{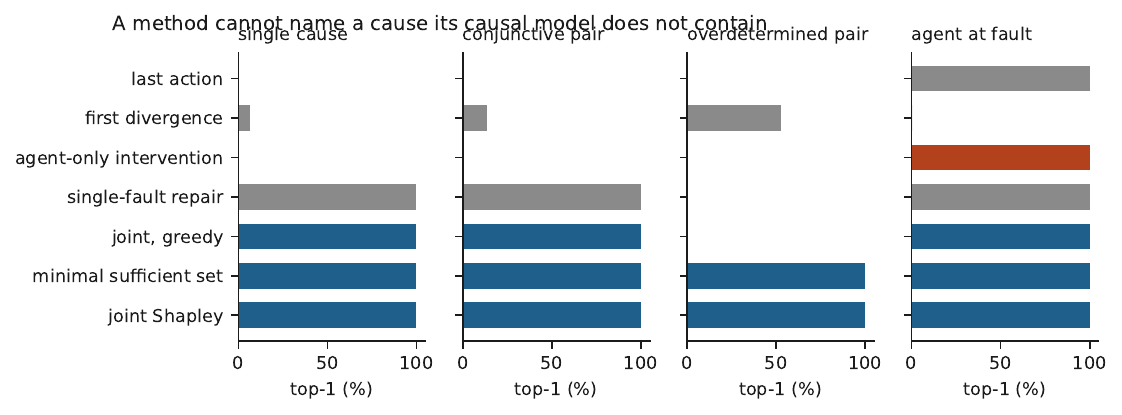}
\caption{Top-1 identification by stratum. Agent-only intervention is correct
exactly when the agent is the cause.}
\label{fig:accuracy}
\end{figure}

\subsection{Naming the cause}

Table~\ref{tab:accuracy} and Figure~\ref{fig:accuracy} give identification
accuracy, and one row of it is not an experimental result at all. Agent-only
intervention identifies the true cause on \agentOnlyInfra\%\ of infrastructure
episodes. That figure is zero and had to be: a causal model whose only
intervenable variables are the agent's steps contains no variable naming a
message, so no infrastructure cause is expressible in its output, whatever the
episode. It is a proposition about the method's model, it needs no corpus, and
running one cannot refute it.

What the corpus measures is the consequence, which does not follow from the
proposition and could have come out otherwise. How much money gets misfiled
(\$\agentOnlyDollars, on \agentOnlyMisattr\%\ of infrastructure episodes), how
little repairing the named cause recovers (\agentOnlyRepair\%), how often
interaction defeats one-at-a-time scoring (\nonAdditiveCI), and what the
alternatives cost in replays are all quantities of this environment and this
generator. A benchmark where infrastructure faults were rare or cheap would
give the same proposition and much smaller numbers.

The method's overall verdict accuracy, \agentOnlyVerdict\%, is the base rate
of episodes in which the agent happens to be at fault. It is not that the
method performs poorly. It answers a different question correctly.

Single-fault repair is exact on the single and conjunctive strata and scores
\faultSingleOverdet\%\ on the overdetermined one, which is the arithmetic:
when either message alone is enough, repairing one changes nothing and every
candidate scores zero. Minimal sufficient sets and joint Shapley are exact on
all four.

First divergence behaves differently on that stratum and the reason is worth a
sentence. It names the earliest message that arrived late or not at all, and
on overdetermined episodes it lands on a true cause about half the time. The
episodes carry distractor faults verified harmless alongside the causal set,
and a distractor is as likely to be early as a cause is, so the earliest
anomaly is frequently not one of the two sufficient messages. The rule is not
reasoning about sufficiency; it is ordering by arrival and taking the first.

\subsection{Splitting the bill}

\begin{table}[t]
\centering
\caption{Allocation between agent and infrastructure. Dollars charged to the
agent are infrastructure-caused loss filed against a decision.}
\label{tab:allocation}
\footnotesize\setlength{\tabcolsep}{3pt}
\begin{tabular}{lrrr}
\toprule
& allocation & misattributed & \$ charged \\
method & error (\%) & (\%) & to agent \\
\midrule
last action & 67.0 & 100 & 114,383.40 \\
first divergence & 33.0 & 0 & 0.00 \\
agent-only intervention & 67.0 & 100 & 114,383.40 \\
single-fault repair & 0.0 & 0 & 0.00 \\
\bfseries joint, greedy & 0.0 & 0 & 0.00 \\
\bfseries minimal sufficient set & 0.0 & 0 & 0.00 \\
\bfseries joint Shapley & 0.0 & 0 & 0.00 \\
\bottomrule
\end{tabular}
\end{table}

Table~\ref{tab:allocation} compares each method's split against the planted
truth. Agent-only intervention misfiles \agentOnlyMisattr\%\ of infrastructure
episodes and charges \$\agentOnlyDollars{} of infrastructure-caused loss to the
agent.

The methods that intervene on faults all reach zero allocation error, and that
deserves a caveat. Once a method can intervene on the
fault process at all, the agent/infrastructure split follows from the identity
in Section~\ref{sec:split} and is exact by construction. The hard part is
\emph{which} message, not \emph{which side}. Table~\ref{tab:allocation} shows
what having the variable at all is worth; Table~\ref{tab:accuracy} shows what
distinguishes the methods that have it.

\subsection{Acting on the attribution}

\begin{table*}[t]
\centering
\caption{Repairing what each method named, and what it cost in episode
replays.}
\label{tab:repair}
\footnotesize\setlength{\tabcolsep}{3pt}
\begin{tabular}{lrrr}
\toprule
& loss removed & fully fixed & replays \\
method & (\%) & (\%) & per ep. \\
\midrule
last action & 0.0 & 0.0 & 1.0 \\
first divergence & 12.5 & 6.7 & 1.0 \\
agent-only intervention & 0.0 & 0.0 & 70.5 \\
single-fault repair & 82.4 & 81.1 & 8.4 \\
\bfseries joint, greedy & 82.4 & 81.1 & 78.0 \\
\bfseries minimal sufficient set & 100.0 & 100.0 & 2.3 \\
\bfseries joint Shapley & 100.0 & 100.0 & 280.1 \\
\bottomrule
\end{tabular}
\end{table*}

\begin{figure}[t]
\centering
\includegraphics[width=\columnwidth]{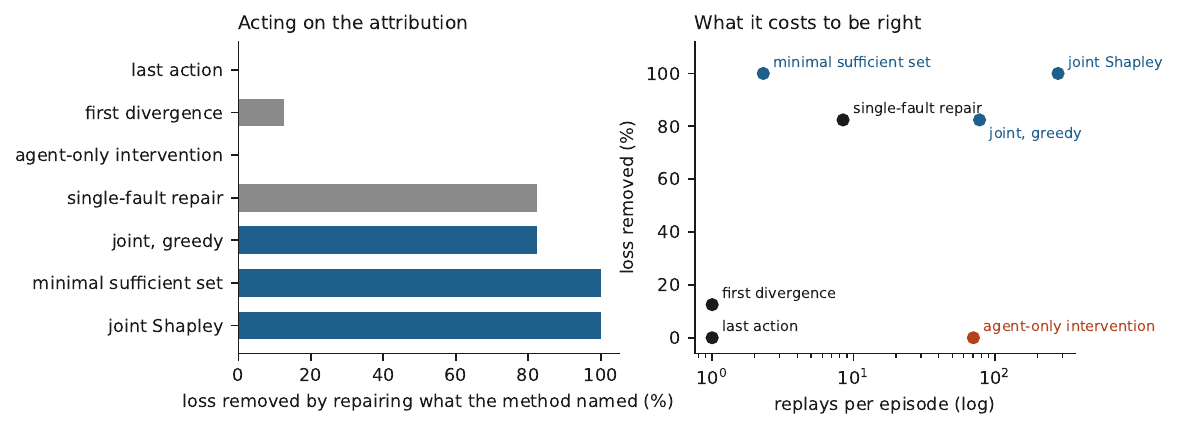}
\caption{Left: loss removed by repairing the named cause. Right: the same
against replay cost. Minimal sufficient sets sit at the top-left corner.}
\label{fig:repair}
\end{figure}

An attribution is worth something if repairing what it names removes the loss.
Of \$\dollarsAvailable{} of infrastructure-caused loss, repairing what agent-only intervention names recovers \agentOnlyRepair\%. It
names agent steps, and an agent step is not a thing an operator can repair. Single-fault
repair recovers \faultSingleRepair\%. Minimal sufficient sets recover
\minSuffRepair\%.

Cost is counted in replays rather than seconds. A replay count is a property
of the method and reproduces exactly; wall-clock time is a property of the
machine, and ours moved threefold between runs that changed no code. Minimal
sufficient sets need \minSuffReplays{} replays per episode; joint Shapley
needs \shapleyReplays.
Shapley buys the full allocation across messages, which a minimal set does not
provide; when the question is only ``what do we fix'', the cheap method is also
the right one.

\subsection{How often interactions matter}

Across \nonAdditiveCI{} of episodes with infrastructure loss, repairing each
message on its own does not account for the whole infrastructure effect. The
mean gap is \$\interactionGap. On \overdetCI{} every single-repair score is
zero while the total is not. We report 95\% percentile intervals from \ciReplicates{} task-clustered
bootstrap replicates over \ciTasks{} tasks covering \ciEpisodes{} episodes,
resampling tasks rather than episodes because episodes from one task share its
amount, archetype and fault draw. The algebraic identity needs no such interval and is not given one;
these are estimates of how often a phenomenon occurs in a generated
population, which is a different kind of claim.

Interaction is not a corner case here. It is roughly a quarter of the
episodes, and the interval says the sample constrains that to between a
quarter and a third.

\subsection{One episode, end to end}

A concrete case makes the difference legible. Take a \texttt{lost\_settlement}
task worth \$\exOrder{} run under the ReAct loop. The capture settles, the
settlement message to the processor is dropped, and \exInstances{} fault
instances are present in all. The policy polls the processor, never sees a
settlement, escalates, and the merchant ends \$\exLoss{} behind the
reference.

The three-way split assigns the whole \$\exLoss{} to infrastructure: in a
repaired world this policy solves the task exactly, so its policy differential
and reference-policy residual are both zero. Shapley then divides that \$\exLoss{} between two
messages at \$\exShare{} each, the dropped settlement to the processor, and a
reordering at the bank feed, and gives zero to the other \exZeroed{}. Neither fault is sufficient by itself to create the loss; repairing either one
alone is sufficient to remove it. That is why they split the credit evenly
instead of one taking all of it.

Agent-only intervention, run on the same episode, reports that the agent should
have shipped at step~0 and assigns \$\exAgentOnly{} to that decision. The recommendation
is not wrong as advice, shipping would indeed have been better, but as an
explanation it inverts the case. At step~0 the agent had no evidence that the
money had arrived, and the reason it had none is the first of the two messages
Shapley named.

\subsection{Where the money actually goes}

\begin{table}[t]
\centering
\caption{The split applied to the whole corpus, without planting.}
\label{tab:landscape}
\footnotesize\setlength{\tabcolsep}{3pt}
\begin{tabular}{lrrrr}
\toprule
& mean loss & infra. & policy & residual \\
policy & (\$) & (\%) & (\%) & (\%) \\
\midrule
optimistic & 166.60 & -6.8 & 88.0 & 18.8 \\
rule based & 213.63 & 75.1 & 10.2 & 14.7 \\
react & 134.15 & 59.3 & 17.3 & 23.4 \\
transactional & 31.37 & 0.0 & 0.0 & 100.0 \\
\bottomrule
\end{tabular}
\end{table}

\begin{figure}[t]
\centering
\includegraphics[width=\columnwidth]{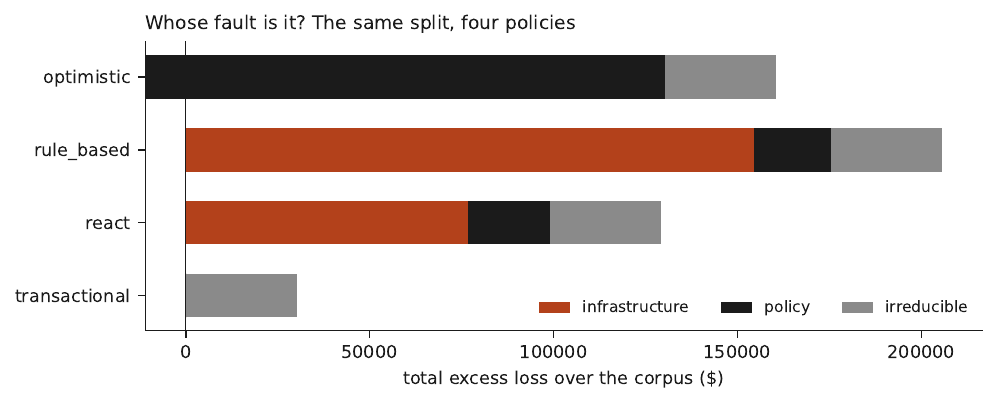}
\caption{The same decomposition for four policies over the whole corpus.}
\label{fig:landscape}
\end{figure}

Applied without planting (Table~\ref{tab:landscape},
Figure~\ref{fig:landscape}), the split says different things about different
policies. For the rule-based procedure, \ruleInfra\%\ of its loss is the
environment's doing. For the ReAct loop, \reactInfra\%\ is infrastructure and
\reactPolicy\%\ is its own. For the finality-gated runtime the infrastructure effect and the policy
differential are both zero, and \gateIrreducible\%\ of its \$\gateMeanLoss{}
mean loss is the reference-policy residual. It loses only what no
unprivileged policy can avoid, which the benchmark's own results reach by a
different route.

The interesting row is the optimistic policy, which ships as soon as the ledger
shows any activity. Its infrastructure effect is \optInfra\%, and it is \emph{negative}.
Faults help it, on net. A policy that acts on the first sign of payment is
occasionally saved by a message going missing, because the message it did not
receive is the one it would have acted on. We would not have predicted the sign
and do not present it as a general result; it is a reminder that ``the
infrastructure was at fault'' is a claim with a direction, and that a method
which can only add blame to the agent cannot represent it at all.

\subsection{Which situations are the environment's fault}

\begin{table}[t]
\centering
\caption{Where responsibility sits, by archetype, pooled over the three non-
gated policies.}
\label{tab:archetype}
\footnotesize\setlength{\tabcolsep}{3pt}
\begin{tabular}{lrrr}
\toprule
archetype & infra. (\%) & policy (\%) & residual (\%) \\
\midrule
stale processor & 99.6 & 0.4 & 0.0 \\
pending settle & 99.6 & 0.4 & 0.0 \\
lost settlement & 99.4 & 0.6 & 0.0 \\
late chargeback & 99.0 & 1.0 & 0.0 \\
partial ledger & 94.8 & 5.2 & 0.0 \\
duplicate settlement & 89.6 & 10.4 & 0.0 \\
pending fail & 0.7 & 94.2 & 5.1 \\
unresolvable & -4.9 & 10.0 & 94.9 \\
pending fail hard & -85.6 & 184.4 & 1.2 \\
\bottomrule
\end{tabular}
\end{table}

Table~\ref{tab:archetype} breaks the same decomposition down by situation. Its
values are the three signed terms as percentages of each archetype's total
excess loss, so a term can be negative and can exceed 100\% when another runs
the other way; they are not shares. It lines up with what the benchmark says
about itself from the other direction.
Infrastructure dominates on the archetypes built around a message going missing,
arriving stale, or posting twice. These are situations in which a competent
policy is defeated by its inputs. The reference-policy residual concentrates on the
cases whose outcome lands after the ship-by deadline, where repaired plumbing
would not help because the fact itself has not happened yet. The practical
reading is that these two groups call for different remedies: the first is
fixed by making delivery reliable or by reading an authoritative channel, and
the second is not fixable by any amount of engineering on the observation path.

\section{Discussion}

\textbf{The variable set is the design decision.} Nothing about CAR-style
intervention is wrong. Given a variable, it estimates that variable's effect
carefully. The result here is about what happens when a variable is missing:
the loss does not go unexplained, it gets attributed to whatever variables
remain. In an agent-analysis tool those are all agent steps, so the tool is
biased toward blaming the agent by construction, not by error.

\textbf{Exactness beats estimation where it is available.} The three-way split
needs no confidence interval because it telescopes. The Shapley step needs one
only above thirteen messages, and even then the sampled estimate is within
\shapleyError\%. Deterministic replay is what buys this, and it is available in
any environment whose fault process is seeded and reproducible.

\textbf{What determinism buys and what it costs.} Every intervention here is
a single replay with a known answer, not a Monte Carlo estimate over a
stochastic policy. That is why the split needs no confidence interval and why
overdetermination can be detected exactly rather than inferred from overlapping
error bars. The price is that our agent-step intervention substitutes an action
rather than resampling one, and those coincide only when the policy is
deterministic. For a stochastic agent the two come apart, and the published
methods we compare against are built for that case; how the comparison moves
there is an open question we cannot settle from inside this environment.

\textbf{Cheap methods can be exact.} Minimal sufficient repair is the cheapest
method in Table~\ref{tab:repair} that is exact on every planted stratum, at
\minSuffReplays{} replays, and the \searchCap-message cap never bound on any
episode searched. It does not win on accuracy per replay taken as a ratio:
first divergence answers with one replay and is right about the side often
enough to beat it on that measure, while being wrong about which message
whenever it matters. The expensive method is worth its cost only when the
full per-message allocation is the deliverable.

\section{Limitations}
\label{sec:limits}

\textbf{No language model was evaluated.} The policies are deterministic
procedures. This is a real gain for the causal analysis, an intervention is
exact, and there is no sampling noise to average away, and a real loss for
external validity. Whether these conclusions hold when the agent is
stochastic, and whether a resampling intervention behaves like our substitution
intervention, is untested here. The environment's model adapter is written and
tested against recorded transcripts but has never called a model.

\textbf{One environment.} Everything is measured inside FinalityBench. Its
faults are delivery-level, so infrastructure harms an agent exclusively by
corrupting what it can see; an environment where infrastructure moves money
directly would need a wider variable set than ours.

\textbf{Overdetermination is rare in nature.} It is \overdetRate\%\ of episodes
with infrastructure loss, and the stratum has \nOverdetermined{} episodes pooled
across \nPolicies{} policies. The conclusion that single-effect scoring fails
there is arithmetic rather than statistical, but the rate at which it costs
anything in practice is estimated from a small sample.

\textbf{Agent-step alternatives are a fixed menu.} Seven dispositions an
operator would recognise, not the whole action space. A better alternative
outside the menu would raise the measured effect of a step, which would make
agent-only attribution look better, not worse.

\textbf{Which policy counts as the reference is a choice.} We use the
benchmark's strongest implementable policy, and that choice fixes the
residual: a stronger reference would move loss out of the residual and into
the policy differential for every subject.

\section{Conclusion}

Given an environment whose fault process can be replayed, the question ``was
this the agent's fault or the system's?'' has an exact answer, and it is cheap
to compute. The obstacle is not statistical. It is that a causal model
containing only agent steps has to answer ``the agent'' every time, and will
do so confidently: on our planted episodes such a baseline misfiles every
infrastructure failure and charges \$\agentOnlyDollars{} to decisions, and
repairing what it names recovers none of it.

Two things follow for anyone building this. Expose environment events as
intervenable variables, because a cause that cannot be named cannot be found,
and the cost of omitting them is measurable rather than theoretical. And when
the goal is operational remediation rather than complete allocation, use
minimal sufficient repair: it recovers \minSuffRepair\%\ of the available loss
at \minSuffReplays{} replays per episode, against \shapleyReplays{} for the
full per-message allocation, which may be unnecessary when the operational
goal is only to identify a sufficient repair.

\section*{Artifact availability}

Code, the planted-episode generator and every result file are at
\url{https://github.com/abhisheksharma2411/causalloss-fin}. It is archived at DOI \texttt{10.5281/zenodo.22893020}. It consumes the released
FinalityBench artifact~\cite{finalitybench} rather than reimplementing it, and
each result file records the revision it ran against. \texttt{make reproduce}
runs the suite end to end from a clean tree.

\end{document}